\documentclass[10pt,twocolumn,letterpaper]{article}
\PassOptionsToPackage{table}{xcolor}
\newcommand{\mname}{Points-to-3D\xspace}
\usepackage[dvipsnames]{xcolor}

\usepackage[pagenumbers]{cvpr} 

\definecolor{cvprblue}{rgb}{0.21,0.49,0.74}
\usepackage[pagebackref,breaklinks,colorlinks,allcolors=cvprblue]{hyperref}

\def\paperID{8659} 
\def\confName{CVPR}
\def\confYear{2026}

\title{Points-to-3D: Structure-Aware 3D Generation with Point Cloud Priors}

\author{
Jiatong Xia\thanks{Jiatong Xia and Zicheng Duan equally contributed to this work.},
~~~
Zicheng Duan\footnotemark[1],
~~~
Anton van den Hengel,
~~~
Lingqiao Liu\thanks{Corresponding author, e-mail: $ \sf lingqiao.liu@adelaide.edu.au $}
\\[0.125cm]
 Australian Institute for Machine Learning, University of Adelaide, Australia
\\[0.1cm]
}

\begin{document}

\twocolumn[{%
\centering
\renewcommand\twocolumn[1][]{#1}%
\maketitle
    \captionsetup{type=figure}
    \vspace{-7mm}
    \includegraphics[width=0.95\textwidth]{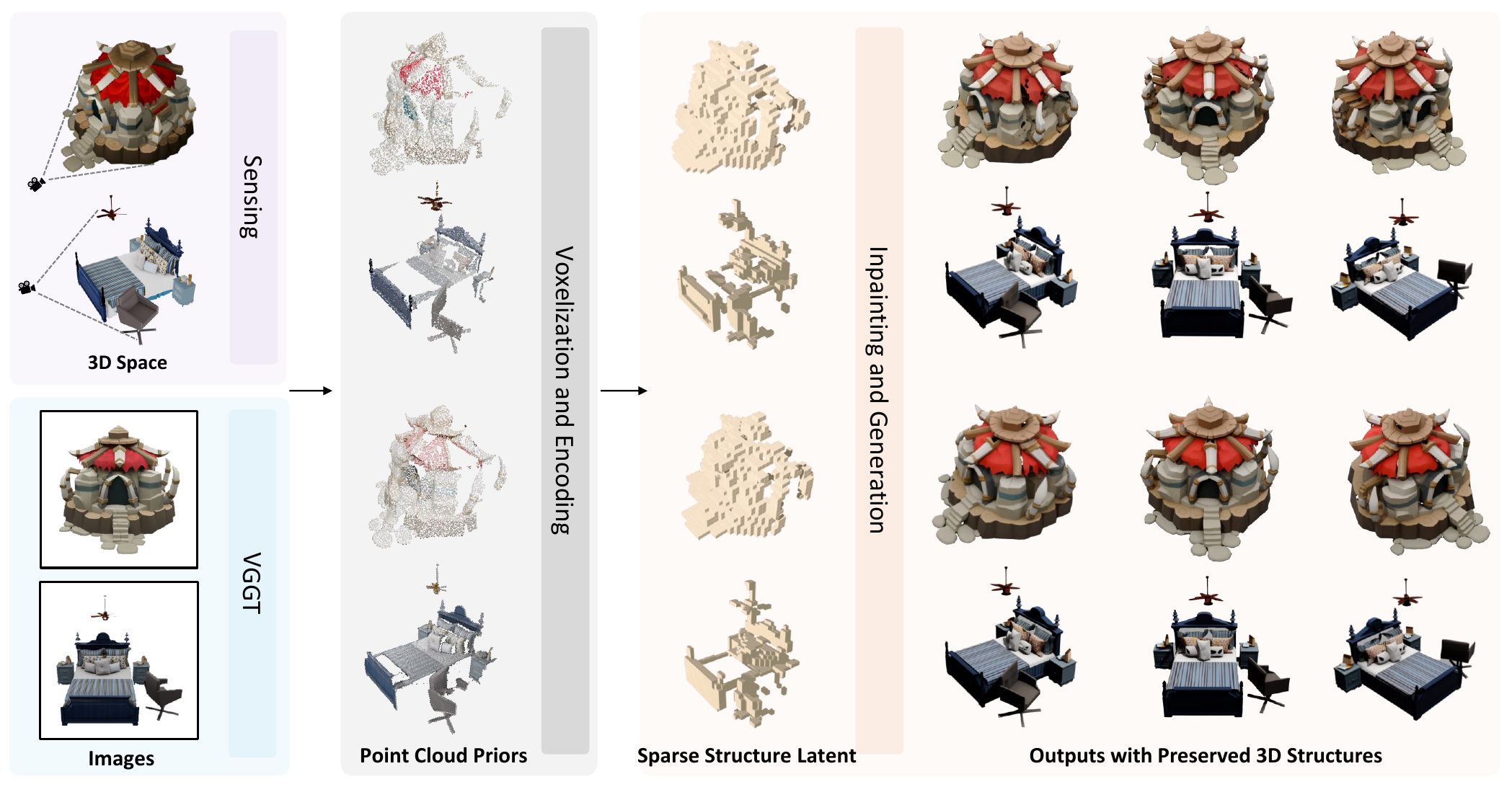} 
    \vspace{-2mm}
    \hfill\caption{We introduce explicit 3D point cloud priors into 3D generation framework, given a pre-existing point cloud or a feed-forward point cloud prediction from image input, our model generates high-quality 3D assets that faithfully preserve the observed structure while plausibly completing unobserved regions with coherent geometry.}
    \label{fig:teaser}
    \hfill 
}]

\renewcommand{\thefootnote}{\fnsymbol{footnote}}
\footnotetext[1]{Jiatong Xia and Zicheng Duan equally contributed to this work.}
\footnotetext[2]{Corresponding author, e-mail: lingqiao.liu@adelaide.edu.au}

\begin{abstract}
Recent progress in 3D generation has been driven largely by models conditioned on images or text, while readily available 3D priors are still underused. In many real-world scenarios, the visible-region point cloud are easy to obtain—from active sensors such as LiDAR or from feed-forward predictors like VGGT—offering explicit geometric constraints that current methods fail to exploit. In this work, we introduce Points-to-3D, a diffusion-based framework that leverages point cloud priors for geometry-controllable 3D asset and scene generation. Built on a latent 3D diffusion model TRELLIS, Points-to-3D first replaces pure-noise sparse structure latent initialization with a point cloud priors tailored input formulation.A structure inpainting network, trained within the TRELLIS framework on task-specific data designed to learn global structural inpainting, is then used for inference with a staged sampling strategy (structural inpainting followed by boundary refinement), completing the global geometry while preserving the visible regions of the input priors.In practice, Points-to-3D can take either accurate point-cloud priors or VGGT-estimated point clouds from single images as input. Experiments on both objects and scene scenarios consistently demonstrate superior performance over state-of-the-art baselines in terms of rendering quality and geometric fidelity, highlighting the effectiveness of explicitly embedding point-cloud priors for achieving more accurate and structurally controllable 3D generation.
Project page: \href{https://jiatongxia.github.io/points2-3D/}{https://jiatongxia.github.io/points2-3D/}
\end{abstract}    
\section{Introduction}
\label{sec:intro}

Advances in 3D generation now allow models to synthesize realistic and diverse 3D assets from single-view images or text prompts. These “foundation” 3D models~\cite{lan2024ga,zhang2024clay,yao2025cast,huang2025midi,xiang2025structured} can produce 3D assets across broad categories, supporting applications in content creation and virtual environments. However, conditioning solely on 2D images or text provides limited geometric controllability: while the output may appear plausible, the model lacks any mechanism to respect real 3D measurements. In practice, partial point clouds from sensors or image-based predictors provide reliable geometry for visible regions, yet current 3D generative pipelines make little use of this readily available structural information.

In this work, we address this gap by enabling geometry-controllable 3D generation driven by point cloud priors. We focus on the setting where a visible-region point cloud—captured or predicted—is treated as a hard structural constraint, requiring the generated asset to align with observed geometry while plausibly completing unobserved parts. Achieving this cannot be done by simply injecting the point cloud as an additional condition; it requires to integrates the structural prior into the latent space itself.

Recent latent 3D diffusion models~\cite{lan2024ga,jiang2024real3d,tang2024lgm}, represented by TRELLIS~\cite{xiang2025structured}, factorize 3D generation into two stages: a coarse structural stage operating on a sparse occupancy representation, followed by a semantic and appearance refinement stage. This paradigm offers an explicit structure latent that could be guided by 3D priors. Yet in their default formulation, these structure latents are initialized purely from Gaussian noise and rely only on text or image embeddings, making them unable to anchor structural generation to actual 3D observations.

To overcome this limitation, we introduce Points-to-3D, a point-cloud–guided 3D generation framework that re-defines how the structural latent is initialized and completed. Instead of starting from pure noise, we voxelize the visible point cloud and encode it with the TRELLIS sparse structure VAE to obtain a partially observed latent that directly reflects the measured geometry. Regions supported by observations are preserved as fixed constraints, while unobserved regions remain free to be synthesized. A mask-aware inpainting network completes this mixed latent, enabling the model to generate coherent structures that respect real 3D measurements while plausibly filling missing areas.

To support this formulation, we construct a visibility-aware training pipeline that first produces realistic partial–complete structure pairs from ground-truth assets, these pairs supervise the inpainting model to generate geometric cues from visible regions to invisible ones while maintaining consistency with the input point cloud. During inference, Points-to-3D adopts a lightweight two-stage procedure: it first establishes a globally consistent structure under visibility constraints, and then performs a brief refinement step to enhance boundary quality without disturbing anchored geometry. This design enables controllable and structurally faithful 3D generation from both sensor-captured and image-predicted point clouds.

We evaluate Points-to-3D on object-level (Toys4K~\cite{stojanov21toys4k}) and scene-level (3D-FRONT~\cite{fu20213dfront}) benchmarks. Across all settings, our method consistently outperforms TRELLIS~\cite{xiang2025structured} and other baselines in rendered-view quality and geometric fidelity. Gains are especially significant in regions covered by point-cloud priors, where Points-to-3D achieves near-perfect alignment while maintaining realistic completions in unseen areas. Furthermore, combining our point-cloud–anchored structure generation with text conditioning enables controllable text-to-3D generation guided by concrete 3D measurements.
\section{Related Work}
\label{sec:related}

\subsection{3D Modeling}
Recovering the 3D model of specific scenes or objects is a fundamental problem in computer vision and graphics. Classical 3D reconstruction leverages multi-images to recover geometry, including Structure-from-Motion (SfM)~\cite{schonberger2016structure}, Multi-View Stereo (MVS)~\cite{yao2018mvsnet,yao2019recurrent,yu2020fast}, SDF-based approaches~\cite{Park_2019_deepsdf,sitzmann2020implicit,zhang2021learning}, e.t.c. Radiance-field models like Neural Radiance Fields (NeRF)~\cite{mildenhall2020nerf,barron2021mip,neff2021donerf,kurz2022adanerf,chen2022tensorf,barron2023zipnerf,xia2025enhancing,kerr2023lerf} and 3D Gaussian Splatting~\cite{kerbl3Dgaussians,Huang2DGS2024,li2024dngaussian,zhang2025mega,jiang2024gaussianshader,liu2024dynamics,Yu2024MipSplatting} further enable high-fidelity reconstruction and novel-view synthesis after scene-specific optimization, and recent feed-forward variants~\cite{chen2021mvsnerf,chen2024mvsplat,chen2024mvsplat360} reducing the per-scene cost. DUSt3R-related feed-forward reconstruction methods~\cite{wang2024dust3r,leroy2024mast3r,zhang2025monst3r,yang2025fast3r,xia2025TID3R,tang2025mv}, exemplified by VGGT~\cite{wang2025vggt}, predict per-pixel point cloud and implicitly handle camera poses, achieving strong performance even with a single image. However, recovering 3D assets from only one image is still beyond the capabilities of reconstruction methods. 3D generative models~\cite{xiang2025structured,lan2024ga,jiang2024real3d,tang2024lgm, li2025triposg,yang2024hunyuan3d,lai2025latticedemocratizehighfidelity3d,TripoSR2024,zhang20233dshape2vecset} effectively address this scenario: conditioned on a single reference image or even text prompt, they can synthesize plausible 3D assets aligned with the reference content.

\subsection{3D Generative Models}
Prior 3D generation relies on GANs~\cite{chan2022efficient,schwarz2020graf,gao2022get3d} produce convincing results, yet their instability restricts scalability and output diversity. models~\cite{ho2020denoising,song2020denoising,lipman2022flow}, starting with 2D generation~\cite{ldm, sdxl, wan}, diffusion-based methods rapidly expanding across a spectrum of 3D representations~\cite{lan2024ga,jiang2024real3d,tang2024lgm,liu2024one++,shi2023zero123++,shi2023mvdream,Szymanowicz_2025_ICCV,zhang2024clay,yao2025cast,huang2025midi,boss2025sf3d,huang2025spar3d}. Recently, TRELLIS~\cite{xiang2025structured} introduces a novel latent representation that enables decoding into versatile 3D output formats, demonstrating strong quality, versatility, and editability, and offering a superior paradigm and framework for 3D generation. Subsequent works~\cite{li2025voxhammer,wu2025amodal3r,yang2025omnipart,meng2025scenegen} have leveraged TRELLIS to implement a wide range of practical applications. Nevertheless, most existing improvements focus on enhancing performance at the reference-conditioning level, while directly embedding 3D priors into latent initialization to enable more reliable generation still remains largely unexplored in 3D generation.

\subsection{Point Cloud Priors}
Incorporating 3D priors to assist downstream tasks has proven to be a effective strategy. Where point cloud stand out as one of the most practical and informative representations. Leveraging point cloud priors has advanced a wide range of 3D perception tasks~\cite{Shi_2019_CVPR, zhao2021point,wu2024point} as well as reconstruction tasks~\cite{kangle2021dsnerf,roessle2022dense}. In particular, visible-region point clouds are easy to obtain from diverse sources, including active sensors such as LiDAR and structured-light depth cameras—now widely accessible even on mobile devices—or from reconstruction approaches such as VGGT~\cite{wang2025vggt}. Integrating these easily obtainable point cloud into 3D generative models offers promising potential for explicit geometry control and accurate modeling of complex multi-object scenes. This work seeks to establish a simple yet effective paradigm for incorporating point cloud priors into a diffusion-based 3D generation framework.

\subsection{Inpainting}
Inpainting is a common paradigm for completing missing content while preserving observed structures. In 2D vision, diffusion-based inpainting methods~\cite{sdedit, brushnet, anydoor, objectmover} use spatial masks to guide the synthesis of occluded regions, achieving coherent and controllable image completion. Similar ideas appear in 3D completion~\cite{chu2024diffcomplete,cheng2023sdfusion,kasten2024point,galvis2024sc} where partial scans are used to infer full geometry, but such approaches usually operate as separate completion modules rather than within a generative framework. TRELLIS, however, has the potential to perform inpainting directly within its sparse structured latent spaces, enabling improved global coherence without the need for external modules. Building on this perspective, we formulate point-cloud–conditioned 3D generation as a latent inpainting problem, allowing observed geometry to be embedded and completed naturally within the generative process without relying on auxiliary completion components.

\section{Method}
\label{sec:method}

\begin{figure*}
\setlength{\abovecaptionskip}{-0.5pt}
    \setlength{\belowcaptionskip}{-5pt}
    \centering
    \includegraphics[width=1\linewidth]{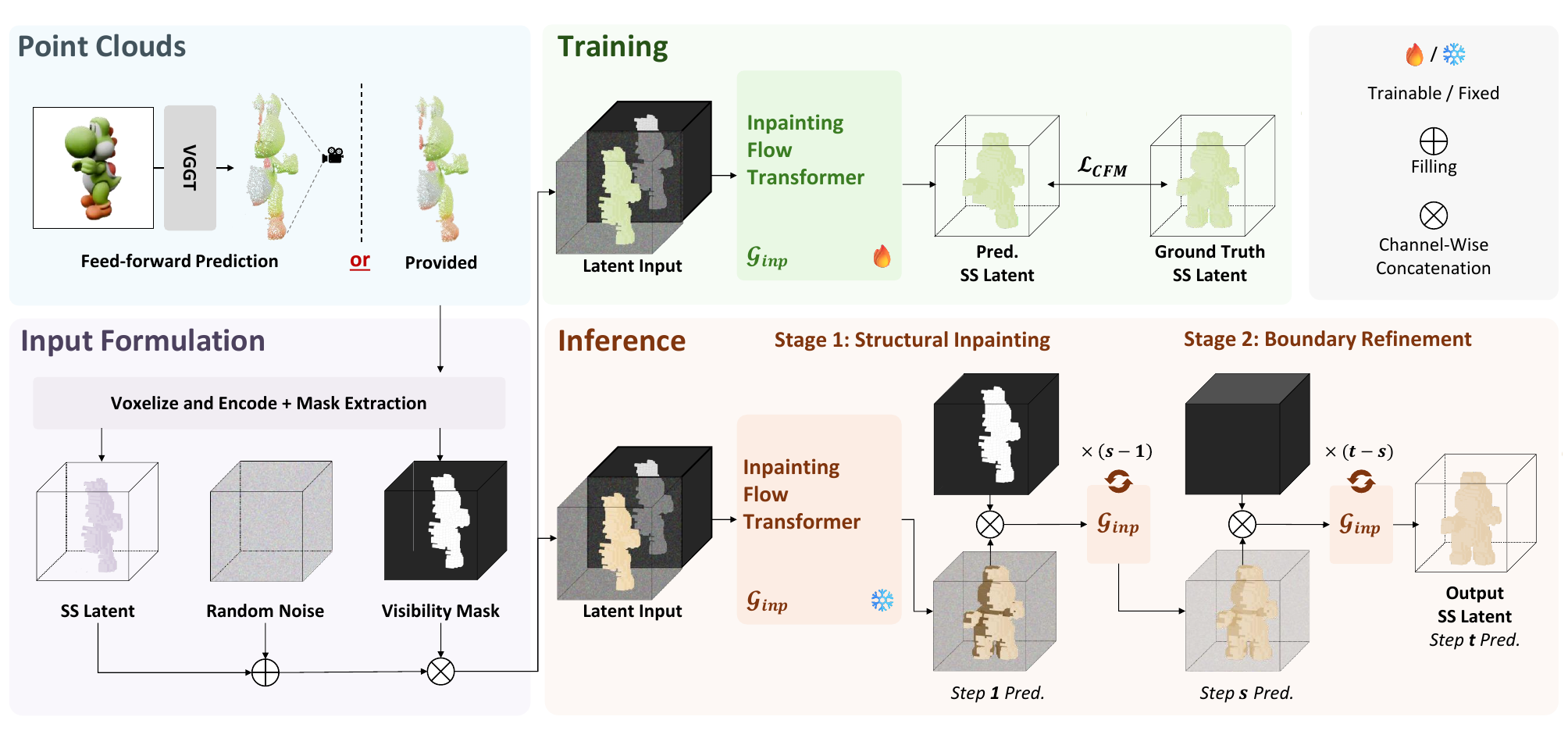}
    \caption{\textbf{Overall framework.} Given point cloud priors—either pre-existing or predicted by VGGT from input image—we first voxelize and VAE-encode it to obtain an SS latent, where the empty regions are filled with random noise and concatenated with an extracted mask to form the input paradigm for our model. During training, the input \textcolor{ForestGreen}{training data} is fed into our inpainting flow transformer $\mathcal{G}_{inp}$, which is optimized via a conditional flow matching loss. During inference, the input \textcolor{Orange}{test data} is processed by the trained $\mathcal{G}_{inp}$ through a two-stage sampling procedure: (1) a structural inpainting stage with $s$ sampling steps to inpaint the global structure. And (2) a boundary refinement stage with remaining $(t-s)$ steps to refine the inpainting boundaries, yielding the final output SS latent.} 
    \label{fig:placeholder}
    \label{pipeline}
\end{figure*}

We seek to achieve geometry-controllable 3D generation by conditioning on point clouds, whether captured by real-world sensors or inferred from a single image via feed-forward prediction. This section first outlines TRELLIS, the baseline that underpins our work, then formalizes the problem and introduces our method, detailing the model architecture, training-data construction, and sampling strategy.

\subsection{Preliminaries: TRELLIS}
TRELLIS~\cite{xiang2025structured} is a recently proposed 3D generation model that produces diverse and high-fidelity 3D assets from image or text prompts. Unlike conventional diffusion models that operate directly in voxel or implicit-function space, TRELLIS performs diffusion in a compact latent space specifically designed to encode 3D structure and appearance. This latent space is learned via a pair of variational autoencoders (VAEs) trained to compress and reconstruct 3D assets. The first VAE encodes voxelized 3D features derived from the original asset into a latent representation called the \emph{structured latent} (SLAT), denoted as $\mathbf{z}=\{(\mathbf{z}_i, \mathbf{p}_i)\}_{i=1}^L$, where $\mathbf{z}_i \in \mathbb{R}^{c_\text{slat}}$ is a local feature attached to voxel position $\mathbf{p}_i \in [0,N-1]^3$ with $N=64$. The SLAT $\mathbf{z}$ can be decoded into multiple 3D output formats—Gaussian splats, radiance fields, or meshes—through corresponding decoders, enabling flexible rendering backends. The second VAE $(\mathcal{E}_{s}, \mathcal{D}_{s})$ learns a compact representation of geometry by encoding a binary voxel occupancy grid $\mathbf{M} \in \{0,1\}^{N \times N \times N}$—whose occupied positions correspond to $\{\mathbf{p}_i\}_{i=1}^L$—into a \emph{sparse structure} (SS) latent $\mathbf{q} \in \mathbb{R}^{r \times r \times r \times c_\text{s}}$ (with $r=16$), which can be decoded back into $\mathbf{M}$.

The generation process in TRELLIS proceeds in two stages following a coarse-to-fine paradigm. 
In the \textbf{Structure Generation} stage, a Flow Transformer $\mathcal{G}_{s}$ takes Gaussian noise $\boldsymbol{\epsilon}_\text{s} \sim \mathcal{N}(0,\mathbf{I})$ and a condition embedding $\mathbf{c}$ (from image or text) to sample the SS latent $\mathbf{q}$, which is then decoded by $\mathcal{D}_{s}$ into a binary voxel grid $\mathbf{M}$, defining the asset’s geometric scaffold. In the subsequent \textbf{Structured Latent Generation} stage, a Sparse Flow Transformer $\mathcal{G}_{l}$ takes noise $\boldsymbol{\epsilon}_\text{slat} \sim \mathcal{N}(0,\mathbf{I})$, the voxel positions $\{\mathbf{p}_i\}_{i=1}^L$, and the same condition $\mathbf{c}$ to generate the SLAT $\mathbf{z}$, which is decoded into the final 3D asset with texture and semantics. Overall, TRELLIS establishes a two-level generative hierarchy that first synthesizes a sparse geometric structure and then enriches it with detailed appearance.

While the VAEs in TRELLIS possess the intrinsic ability to encode meaningful 3D geometry, the generative process itself is not conditioned on external 3D information. Our approach, \textbf{Points-to-3D}, leverages this encoded structural capability by directly injecting point-cloud priors into the VAE latent space, thereby grounding the diffusion process to explicit 3D observations.

\subsection{Problem Formulation}
In many real-world settings, we aim to generate 3D assets conditioned on point cloud priors—obtained either via active sensing (e.g., LiDAR) or model prediction (e.g., VGGT). These point cloud typically cover only the visible portion of the scene. In such a case, the goal is to use the visible-region point cloud $\mathbf{P}$ as a prior for geometry-controllable 3D asset generation: preserving the observed foreground structure while completing unobserved regions guided by foreground cues. To this end, we cast the task as inpainting conditioned on $\mathbf{P}$, inferring missing geometry from the surrounding latent context.

Specifically, unlike TRELLIS—which initializes generation from pure noise $\boldsymbol{\epsilon}_\text{s}$—our structural inpainting stage begins by voxelizing the visible point-cloud priors $\mathbf{P}$ into a binary 3D occupancy grid $\mathbf{M'} \in \{0,1\}^{N \times N \times N}$. This voxelized structure is then encoded with the VAE encoder $\mathcal{E}_{s}$ to obtain the initial SS latent $\mathbf{q_\text{vis}}\in\mathbb{R}^{r\times r\times r\times c_\text{s}}$ (with $r=16$), which serves as the generation starting point. Formally:
\begin{align}
    \mathbf{q_\text{vis}} &= \mathcal{E}_{s}(\mathbf{M'}).
\end{align}
To indicate which SS latent regions should be preserved, we derive an occupancy mask $\mathbf{m_\text{s}}\in\mathbb{R}^{r\times r\times r\times c_\text{m}}$ by down-sampling $\mathbf{M'}$ to the latent resolution. Then, we preserve the visible region SS latent with $\mathbf{m_\text{s}}$ and fill the remaining with noise to obtain the inpainting input SS latent $\mathbf{q}_\text{comb}$:
\begin{equation}
    \label{eq: inp1}
    \mathbf{q}_\text{comb}
    = \mathbf{m}_\text{s} \odot \mathbf{q_\text{vis}}
    + (1-\mathbf{m}_\text{s}) \odot \boldsymbol{\epsilon}_\text{s}.
\end{equation}
Ultimately, we aim to build an inpainting model $\mathcal{G}_{inp}$ based on the structure generation model $\mathcal{G}_{s}$ to take $\mathbf{q}_\text{comb}$ as input, and inpaint the final SS latent $\mathbf{q}$, facilitating visible regions geometry-controllable generation.

\subsection{Point Clouds Priors Driven Generative Model}
\textbf{Model design.} As shown in the purple box of \cref{pipeline}, to enforce the inpainting model, $\mathcal{G}_{inp}$, on distinguishing the regions to preserve and generate, we further concatenate the mask $\mathbf{m}_\text{s}$ to the $\mathbf{q}_\text{comb}$ along the channel dimension. This turn $\mathbf{q}_\text{comb}$ to $\mathbf{x}_\text{inp}$:
\begin{equation}
    \label{eq: concat}
    \mathbf{x}_\text{inp}
    = \mathrm{Concat}[\mathbf{q}_\text{comb},\, \mathbf{m}_\text{s}],
    \mathbf{x}_\text{inp}
    \in \mathbb{R}^{r\times r\times r\times (c_\text{s}+c_\text{m})}
\end{equation}
To adapt $\mathcal{G}_{inp}$ with more input channels, we simply replace its input layer inherited from $\mathcal{G}_{s}$ by a newly registered projection layer with channel dimension $(c_\text{s} + c_\text{m})$, and maintain all other network structures unchanged. Then, we fully fine-tune $\mathcal{G}_{inp}$ to learn to inpaint a completed sparse structure latent $\mathbf{q}_\text{pred} \in \mathbb{R}^{r \times r \times r \times c_\text{s}}$ using Conditional Flow Matching loss (CFM) and regard the ground-truth sparse latent $\mathbf{q}_\text{gt}$ as supervision. This can be formulated as:
\begin{equation}
    \mathcal{L}_{CFM}
    = \mathbb{E}_{t,\mathbf{q}_{\text{gt}},\epsilon}
      \left\|
      \mathcal{G}_{inp}(\mathbf{x_\text{inp}},\, t)
      - \left(\epsilon - \mathbf{q_\text{gt}}\right)
      \right\|_2^2
\end{equation}
Note that the condition $c$ and time-dependent noise scheduling for $\textbf{x}_\text{inp}$ are omitted for simplicity.

\begin{figure*}[t]
    \setlength{\belowcaptionskip}{-6pt}
    \centering
    \includegraphics[width=0.82\linewidth]{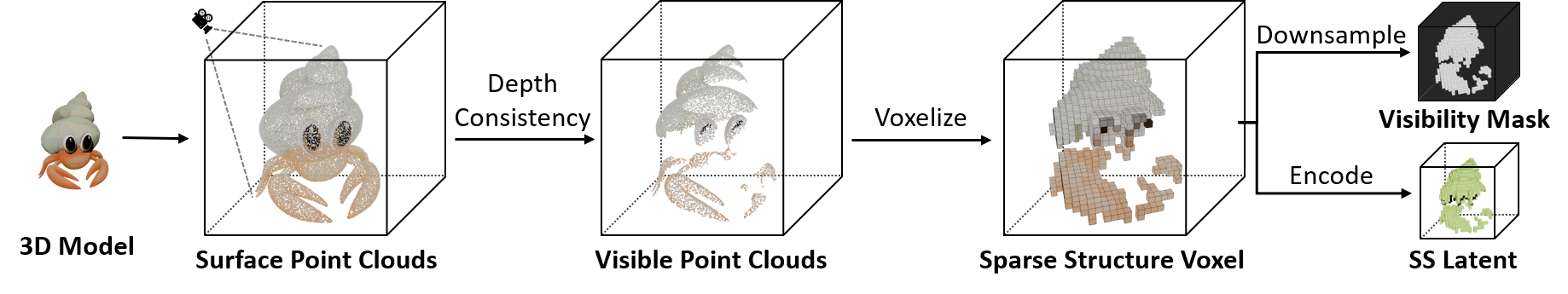}
    \vspace{-8pt}
    \caption{\textbf{Training data processing.} We preserve the visible portion of the complete point cloud and convert it into training inputs.}
    \label{fig:data}
\end{figure*}

\noindent
\textbf{Training data from visible point clouds.} We construct diverse pairs of training data from visible point clouds together with their corresponding ground-truth sparse structure latent to train our model as illustrated in Fig.~\ref{fig:data}. The main challenge lies in accurately obtaining the visible-region point clouds corresponding to the input condition images of each 3D asset. To achieve this, we render the depth map $\mathbf{D}_t$ with height and width as $H$ and $W$ from $T$ viewpoints with the condition images $\mathbf{I}_t$ for each ground-truth 3D asset. For each object, we first uniformly sample $S$ surface points $\hat{\mathbf{P}}=\{\hat{\mathbf{p}}_i=(u_i,v_i,w_i)\}^S_{i=1}$, and given the world-to-camera transformation $\mathbf{T}_t=\left [ \mathbf{R}_t \mid \mathbf{t}_t \right ] $ for the $t$-th view, each point is transformed to the camera as:
\begin{equation}
    \hat{\mathbf{p}}^t_i = \mathbf{R}_t \left ( \hat{\mathbf{p}}_i - \mathbf{t}_t \right ) = \left ( u^t_i,v^t_i,w^t_i \right )^\top
\end{equation}
The corresponding image-plane projection $\mathbf{u}_t \in [1, H] \times [1, W]$ is computed using the intrinsic matrix $\mathbf{K}$. We apply an observation mask $\mathbf{O^t}$ to indicate which point is considered visible in view $t$ if its projected depth $w^t_i$ is consistent with the rendered depth within a tolerance threshold $\tau $:
\begin{equation}
 \mathbf{O}^t_i= \begin{cases}
 1, & \text{ if } \left |\mathbf{D}_t\left ( \mathbf{u}_i\right ) -  w^t_i \right | < \tau , \\
 0, & \text{ otherwise. }
\end{cases}
\end{equation}
The visible point cloud $\mathbf{P}_t$ for view $t$ is thus obtained as $\mathbf{P}_t=\left \{ \hat{\mathbf{p}}^t_i \mid \mathbf{O}^t_i=1  \right \}$. Each $\mathbf{P}_t$ is then voxelized into the sparse structure voxel, which is then encoded and calculated to the SS latent $\mathbf{q}_{\text{comb}}^t$. Simultaneously, the downsampled occupancy mask $\mathbf{m}_\text{s}^t$ is obtained to indicate the visible region of the obtained SS latent. The ground-truth SS latent $\mathbf{q}_\text{gt}$ is derived from the complete 3D structure of the object. Consequently, the samples $\left( \mathbf{q}_{\text{comb}}^t, \mathbf{m}_{\text{s}}^t, \mathbf{I}_t,\mathbf{q}_\text{gt}\right)$ are used to supervise the model $\mathcal{G}_{s}$ (green box in Fig~\ref{pipeline}) to learn structure completion from visible-region priors.

\subsection{Staged Sampling from Point Cloud Priors}
During inference, we split the $t$ step generation into two separate sampling stages, namely the structural inpainting stage and the boundary refinement stage. As illustrated in \cref{pipeline} orange box, the first stage produces a coarse but globally consistent skeleton structure guided by the visible point clouds using inpainting, while the second stage refines the boundary regions that connect newly generated content to the predefined visible areas. Specifically, in each sampling step of the structural inpainting stage, the trained model outputs $\mathbf{q}_\text{pred}$ and reconstructs the inpainting input $\mathbf{x}_\text{inp}$ for the next iteration by concatenating $\mathbf{q}_\text{pred}$ with the visibility mask $\mathbf{m}_\text{s}$ following \cref{eq: concat}. We repeat this process for $s$ steps to obtain a draft skeleton 3D structure that is mostly coherent with the visible point cloud. However, slight inconsistencies and missing details may appear around the boundary regions between generated and predefined visible areas, mainly due to information loss introduced during down-sampling. To address this, we define the latter $(t-s)$ steps as the boundary refinement stage. Here, we replace the visibility mask $\mathbf{m}_\text{s}$ with an all-ones mask $\mathbf{m}_1$, effectively converting inpainting into standard denoising. This allows the model to refine details on either side of the masked or unmasked regions without drastically modifying the existing global geometry, resulting in a fully completed and high-quality sparse structure.

\section{Results}
\label{sec:results}

\begin{table*}[t]
  \centering
  \caption{\textbf{Comparison on single-object generation on Toy4K dataset.} We showcase the performance of our method in two scenarios: one where explicit point cloud priors are provided, and another where point cloud are inferred from condition images using VGGT~\cite{wang2025vggt}. 
  }
  \label{tab:toys4k}
  \resizebox{0.95\linewidth}{!}{
  \begin{tabular}{@{}l|cccc|cccc@{}}
    \toprule
     &\multicolumn{4}{c|}{Rendering}  & \multicolumn{4}{c}{Geometry} \\
      Method  & $\text{PSNR}\uparrow$ & $\text{SSIM}(\%)\uparrow$ &  $\text{LPIPS}\downarrow$ & $\text{DINO}(\%)\downarrow$ & $\text{CD}\downarrow$ & $\text{F-Score}\uparrow$ & $\text{PSNR-N}\uparrow$ & $\text{LPIPS-N}\downarrow$ \\
    \midrule
    GaussianAnything~\cite{lan2024ga}   & 20.08 & 89.31 & 0.183 & 26.74 & 0.084 & 0.513 & 20.99 & 0.199\\
    Real3D~\cite{jiang2024real3d}   & 19.55 & 90.65 & 0.169 & 27.65 & 0.065 & 0.574 & 21.31 & 0.178\\
    LGM~\cite{tang2024lgm}   & 20.55 & 89.98 & 0.181 & 23.45 & 0.075 & 0.487 & 20.04 & 0.202 \\
    VoxHammer~\cite{li2025voxhammer}~(3D Inversion)  & 20.51 & 90.01 & 0.123  & 15.10 & 0.046 & 0.724  & 20.28 & 0.158 \\
    TRELLIS~\cite{xiang2025structured}  & 21.94 & \cellcolor{yellow!30}91.46 &  \cellcolor{yellow!30}0.105 & \cellcolor{yellow!30}7.82 & 0.034 & 0.832 & 23.81 & 0.105\\
    SAM3D~\cite{sam3d_meta2025}  & \cellcolor{yellow!30}22.42 & 91.45 & 0.111 & 8.01 & \cellcolor{yellow!30}0.033 & \cellcolor{yellow!30}0.835 & \cellcolor{yellow!30}23.85 & \cellcolor{yellow!30}0.101\\
    \textbf{\mname} (Ours-VGGT Esti.)  & \cellcolor{orange!30}22.55 & \cellcolor{orange!30}92.09 & \cellcolor{orange!30}0.088  & \cellcolor{orange!30}7.37 & \cellcolor{orange!30}0.024 & \cellcolor{orange!30}0.881 & \cellcolor{orange!30}24.53  & \cellcolor{orange!30}0.085  \\
    \textbf{\mname} (Ours-P.C.Priors)  & \cellcolor{red!30}\textbf{22.91} & \cellcolor{red!30}\textbf{92.83} & \cellcolor{red!30}\textbf{0.070}  & \cellcolor{red!30}\textbf{7.29} & \cellcolor{red!30}\textbf{0.013}  & \cellcolor{red!30}\textbf{0.964} & \cellcolor{red!30}\textbf{27.10}  & \cellcolor{red!30}\textbf{0.053} \\

    \bottomrule
  \end{tabular}}
\end{table*}

\begin{figure*}
\vspace{-2mm}
    \centering
    \setlength{\abovecaptionskip}{-1pt}
    \setlength{\belowcaptionskip}{-6pt}
    \includegraphics[width=1\linewidth]{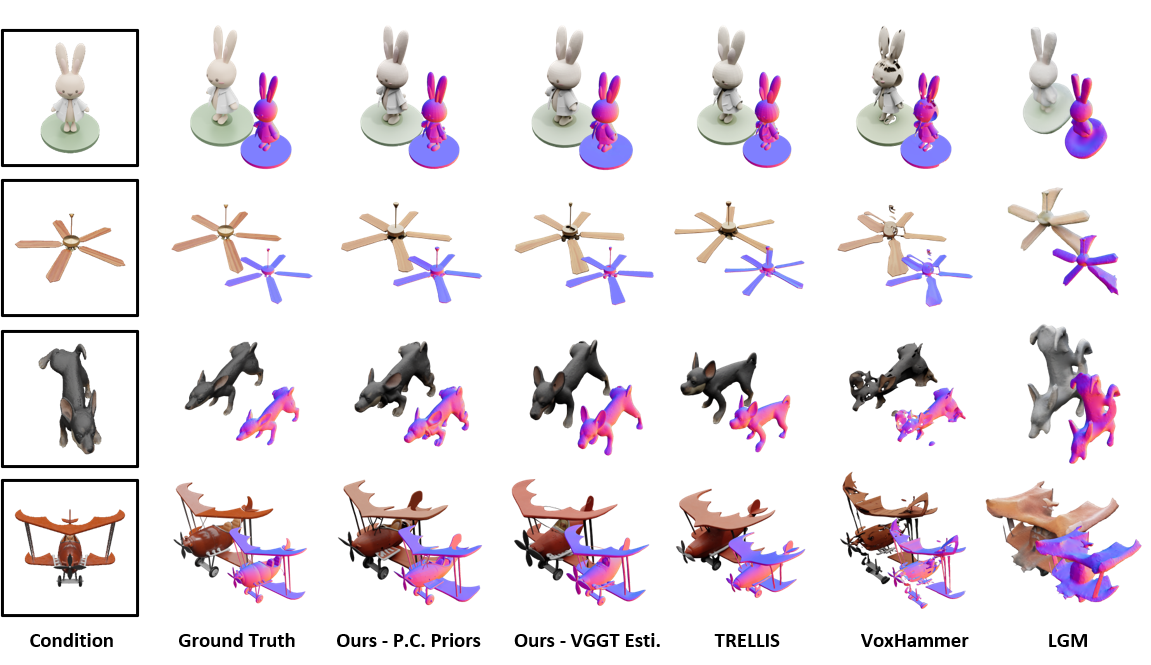}
    \caption{\textbf{Single-object generation on Toys4K.} For the explicit point cloud priors results, we use point cloud extracted strictly from the visible region of input images, whereas the ``VGGT-estimated'' results use point clouds inferred from the condition images by VGGT.}
    \label{fig:toys4k}
\end{figure*}

\subsection{Experiments Setup}

\noindent
\textbf{Datasets.} We train our model on a combination of three datasets:  3D-FUTURE~\cite{fu20203dfuture} dataset, HSSD~\cite{khanna2024hssd} dataset and ABO~\cite{collins2022abo} dataset. During training, we render $T=24$ input views for each object and sample $S=50,000$ point clouds from each object mesh. For each views, we compute the corresponding visible point clouds, which is then process to an initial SS latent used for training. We evaluate our model on two types of test datasets: the single-object dataset Toys4K~\cite{stojanov21toys4k} and the scene dataset 3D-FRONT~\cite{fu20213dfront}. And our method is tested under two settings to cover a broader range of application scenarios. In the first setting, we use the sampled visible point cloud from each view as available input. And in the second setting, where no preprocessed point cloud priors are provided, the test view condition image is fed into VGGT to obtain an estimated point cloud for our model as the initial point cloud priors. We also evaluate our method on several real-world images from the Pix3D~\cite{pix3d} dataset.

\noindent
\textbf{Evaluation metrics.} We evaluate the final generation results in two aspects. For the rendered images of the generated 3D assets, we assess the image quality by comparing with those rendered from the ground-truth 3D asset, and using PSNR, SSIM, LPIPS~\cite{LPIPS}, and DINO~\cite{dinov2} feature similarity as evaluation metrics. For the geometric quality, we employ Chamfer Distance and F-score, as well as the rendered normal maps PSNR and LPIPS as evaluation metrics. For text-to-3D generation evaluation, we use the CLIP~\cite{CLIP} score to measure the consistency between the generated results and the input text prompts.

\noindent
\textbf{Implementation.} We train our model for 20k iterations with a batch size of 8 on 4 Nvidia A100 GPUs, and following the other TRELLIS~\cite{xiang2025structured}'s sparse structure flow transformer's training setting. During inference, we set the $t=50$ sampling steps for the trained Sparse Structure Flow Transformer, allocating $s=25$ steps for structural inpainting and the remaining steps for refinement. For the other comparison methods, we use their official code and settings to reproduce their results. We reproduce the results of the 3D editing method VoxHammer~\cite{li2025voxhammer} to represent the 3D inversion's results. Specifically, we use the structural voxels obtained from the same initial point clouds as in our method to define the ``Unedited Region'' in VoxHammer, and then apply their pipeline to obtain the final generated results.

\begin{table*}[t]
  \centering
  \caption{\textbf{Comparison on scene-level generation on 3D-FRONT dataset.} \mname consistently outperforms state-of-the-art multi-object generation methods across all evaluation metrics.
  }
  \label{tab:3dfront}
  \resizebox{0.95\linewidth}{!}{
  \begin{tabular}{@{}l|cccc|cccc@{}}
    \toprule
    &\multicolumn{4}{c|}{Rendering}  & \multicolumn{4}{c}{Geometry} \\
    Method  & $\text{PSNR}\uparrow$ & $\text{SSIM}(\%)\uparrow$ & $\text{LPIPS}\downarrow$ & $\text{DINO}(\%)\downarrow$ & $\text{CD}\downarrow$ & $\text{F-Score}\uparrow$ & $\text{PSNR-N}\uparrow$ & $\text{LPIPS-N}\downarrow$ \\
    \midrule
    TRELLIS~\cite{xiang2025structured}  & 18.21 & 83.12 & 0.239  & \cellcolor{yellow!30}12.33 & 0.094 & 0.478 & 18.76 & 0.258 \\
    VoxHammer~\cite{li2025voxhammer}~(3D Inversion)  & 19.29 & 84.70 & 0.179  & 18.41 & 0.051 & 0.686 & 20.43 & 0.181  \\
    SceneGen~\cite{meng2025scenegen}  & 18.32 & 83.35 & 0.231  & 14.43 & 0.086 & 0.485 & 19.08 & 0.229  \\
    MIDI~\cite{huang2025midi}  & \cellcolor{yellow!30}19.23 & \cellcolor{yellow!30}85.59 & \cellcolor{yellow!30}0.166  & 14.25 & \cellcolor{yellow!30}0.075 & \cellcolor{yellow!30}0.513 & \cellcolor{yellow!30}20.82 & \cellcolor{yellow!30}0.164 \\
    \textbf{\mname} (Ours-VGGT Esti.)  & \cellcolor{orange!30}20.52 & \cellcolor{orange!30}86.51 & \cellcolor{orange!30}0.152  & \cellcolor{orange!30}8.90 & \cellcolor{orange!30}0.040 & \cellcolor{orange!30}0.743 & \cellcolor{orange!30}20.97  & \cellcolor{orange!30}0.160  \\
    \textbf{\mname} (Ours-P.C.Priors)  & \cellcolor{red!30}\textbf{21.63} & \cellcolor{red!30}\textbf{87.73} & \cellcolor{red!30}\textbf{0.124}  & \cellcolor{red!30}\textbf{8.29} & \cellcolor{red!30}\textbf{0.025} & \cellcolor{red!30}\textbf{0.886} & \cellcolor{red!30}\textbf{22.38} & \cellcolor{red!30}\textbf{0.124}  \\

    \bottomrule
  \end{tabular}}
\end{table*}

\begin{figure*}
    \setlength{\abovecaptionskip}{-1pt}
    \setlength{\belowcaptionskip}{-4pt}
    \centering
    \includegraphics[width=1\linewidth]{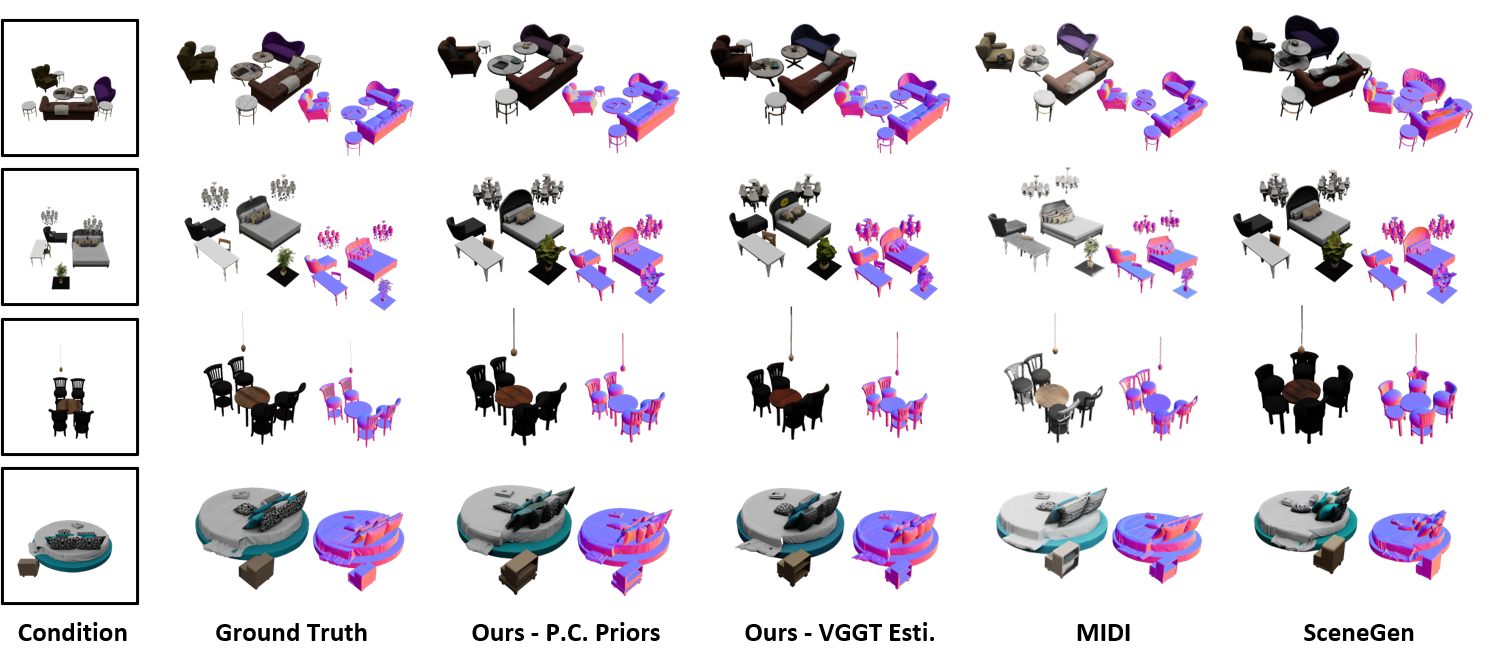}
    \caption{\textbf{Scene-level generation on 3D-FRONT.} The input point cloud priors setting is the same as in Fig.~\ref{fig:toys4k}.}
    \label{fig:3dfront}
\end{figure*}

\begin{table}[t]
  \centering
  \caption{\textbf{Comparison on visible and overall geometry results on Toys4K.} We present the comparison between our method and TRELLIS. For each method, the upper row (O.) shows the overall results, while the lower row (V.) shows the visible region results.
  }
  \label{tab:vis}
  \resizebox{\linewidth}{!}{
  \begin{tabular}{l|cccc}
    \toprule
    Methods  & $\text{CD}\downarrow$ & $\text{F-Score}\uparrow$ & $\text{PSNR-N}\uparrow$ & $\text{LPIPS-N}\downarrow$  \\
    \midrule
    TRELLIS~\cite{xiang2025structured}-O. & 0.034 & 0.832 & 23.81 & 0.105  \\    
    TRELLIS~\cite{xiang2025structured}-V. & 0.032 & 0.854 & 24.77 & 0.093  \\
    SAM3D~\cite{sam3d_meta2025}-O. & 0.033 & 0.835 & 23.85 & 0.101  \\
    SAM3D~\cite{sam3d_meta2025}-V. & 0.031 & 0.841 & 24.81 & 0.090  \\
    \midrule
    \textbf{\mname}-O. & \textbf{0.013} & \textbf{0.964} & \textbf{27.10} & \textbf{0.053}  \\
    \textbf{\mname}-V. & \textbf{0.007} & \textbf{0.998} & \textbf{29.00} & \textbf{0.036}  \\

    \bottomrule
  \end{tabular}}
\end{table}

\subsection{Main Results}

\noindent
\textbf{Single-object generation.}
We first present the results of single-object generation on Toys4K~\cite{stojanov21toys4k}. As shown in Tab.~\ref{tab:toys4k}, our method consistently outperforms existing approaches across all evaluation metrics, whether using existing point cloud priors or VGGT~\cite{wang2025vggt}-predicted point cloud. Notably, in terms of geometric metrics, the results with point cloud priors achieve an F-score of 0.963, demonstrates our approach produces geometry closely approximates the ground-truth structure. The significant improvement in geometry enhances the visual fidelity of the results. As illustrated in Fig.~\ref{fig:toys4k}, our results better match the overall appearance of the ground-truth compared to other methods, and normal maps further highlight the superior geometric quality achieved by our approach. Notably, while VoxHammer adopts the same 3D priors as ours, the image condition fails to provide cues for the missing parts of the 3D priors, making 3D inversion process struggles to complete the unknown regions. SAM3D~\cite{sam3d_meta2025} highlights the value of 3D priors and also leverages point maps, it integrates these priors indirectly through the attention mechanism of the flow transformer block, which—as also stated in their paper—does not support explicit geometric control and exhibits limited ability to enforce precise geometric control compared to our approach. In contrast, our method leverages the trained model’s inpainting capability to fully exploit the existing 3D priors and effectively infer the missing geometry. 

\noindent
\textbf{Scene-level generation.} We evaluate our method on the scene-level generation dataset 3D-FRONT~\cite{fu20213dfront}. As shown in Tab.~\ref{tab:3dfront}, incorporating point cloud priors provides substantial guidance for reconstructing overall geometry in complex scene scenarios, which support our method achieves significant improvements across all evaluation metrics compared to other methods. The rendered images and normal maps in Fig.~\ref{fig:3dfront} further demonstrate that our results better align with the ground-truth scene geometry. Unlike MIDI~\cite{huang2025midi} or SceneGen~\cite{meng2025scenegen}, which implicitly utilize spatial information, our framework explicitly incorporates geometric priors within the architecture, enabling more direct and effective control over 3D geometry, offering a promising solution for generating complex 3D scenes.

\noindent
\textbf{Visible region performance.}
We also highlight the generation results for the visible regions—i.e., the areas covered by our point cloud priors. As shown in Tab.~\ref{tab:vis}, within these visible regions, our generated results achieve an F-Score of 0.998 and a chamfer distance of 0.007, indicating a strong alignment with the ground truth structure. This demonstrates our structure generation pipeline effectively preserves the information provided by the point cloud priors while producing high-quality overall geometry. Compared with others, both in the visible regions and across the entire structure, our approach achieves substantial improvements in geometric fidelity, fulfilling the primary objectives of our work. SAM3D does not achieve improved geometry even within the regions covered by the input pointmap (i.e., the visible areas in the table). While our method injects 3D priors through a more direct and explicit mechanism, enabling effective and reliable geometric controllability, providing current 3D generation frameworks a stronger opportunity to benefit from sensed 3D priors as well as future improvements in feed-forward point-map prediction methods.

\begin{table}[t]
  \centering
  \caption{\textbf{Ablation study.}  We evaluate the number of inpainting steps (Inp.) and refinement steps (Ref.) in our sampling strategy.
  }
  \label{tab:abla}
  \resizebox{0.95\linewidth}{!}{
  \begin{tabular}{ll|cccc}
    \toprule
    Inp. & Ref.  & $\text{CD}\downarrow$ & $\text{F-Score}\uparrow$ & $\text{PSNR-N}\uparrow$ & $\text{LPIPS-N}\downarrow$ \\
    \midrule
    50 & 0 & 0.014 & 0.960 & 25.88 & 0.065 \\
    40 & 10 & 0.013 & 0.962 & 26.49 & 0.059 \\
    30 & 20 & 0.013 & 0.963 & 26.89 & 0.056 \\
    25 & 25 & \textbf{0.013} & \textbf{0.963} & \textbf{27.10} & \textbf{0.053} \\
    20 & 30 & 0.013 & 0.962 & 27.03 & 0.055 \\
    10 & 40 & 0.014 & 0.961 & 26.72 & 0.061 \\
    \bottomrule
  \end{tabular}}
\end{table}

\begin{figure}[h!]
\vspace{-2mm}
    \setlength{\belowcaptionskip}{-3pt}
    \centering
    \includegraphics[width=1\linewidth]{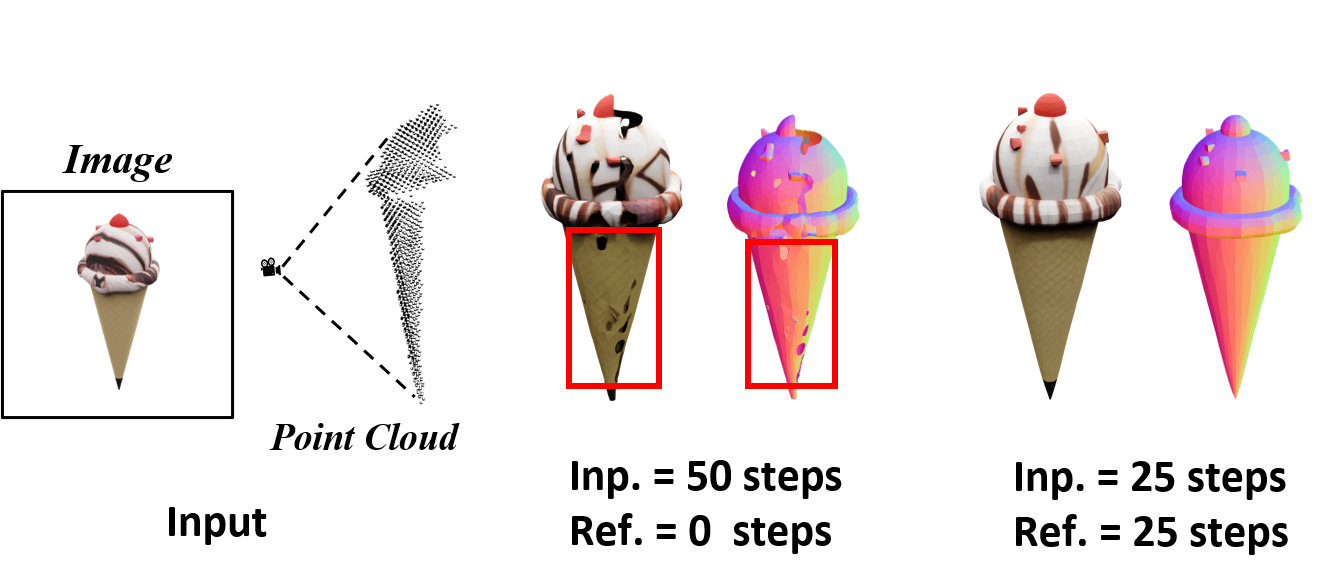}
    \vspace{-7mm}
    \caption{\textbf{Ablation study.} Allocating the full sampling to inpainting (Inp.) results in geometric “holes” along the inpainting edge.}
    \label{fig:abla}
\end{figure}

\subsection{Ablation Studies}
\noindent
\textbf{VGGT point clouds estimation.} When point cloud are not available as input, our method can also leverage the condition image to predict an initial point cloud using feed-forward methods like VGGT~\cite{wang2025vggt}. We evaluate the generation results based on VGGT-estimated point cloud, as shown in Tab.~\ref{tab:toys4k} and Tab.~\ref{tab:3dfront}. Although the generation results with VGGT point cloud exhibit some gap compared to using accurate point cloud priors—this is largely due to the inherent prediction errors of VGGT. Nevertheless, compared to other existing approaches, it consistently achieves substantial improvements in both geometric accuracy and visual fidelity. These highlight the strong robustness and flexibility of our pipeline, with the absence of high-precision priors, our framework can still effectively utilize predicted point cloud form image-only inputs to achieve high-quality geometry generation.

\noindent
\textbf{Staged sampling strategy.}
We propose a staged sampling strategy in our pipeline, which leverages a limited number of last steps with noise to perform global optimization, effectively addressing the “holes” along inpainting boundaries that are otherwise difficult to avoid. We investigate the effect of refinement step allocation through an ablation study. In Tab.\ref{tab:abla}, we present generation results under different allocations of inpainting and refinement steps with the same total sampling steps. When the entire sampling process is allocated to inpainting, the geometric reconstruction suffers from ``holes'' on inpainting edge, as further illustrated in Fig.~\ref{fig:abla}. By setting the sampling schedule to 25 inpainting steps followed by 25 refinement steps, the geometric metrics reach their best performance, and the previously observed ``holes'' are effectively eliminated as in Fig.~\ref{fig:abla}, yielding the overall best generation results.

\subsection{Real-world Input Examples}
\begin{figure}
    \centering
    \vspace{-1mm}
    \includegraphics[width=0.94\linewidth]{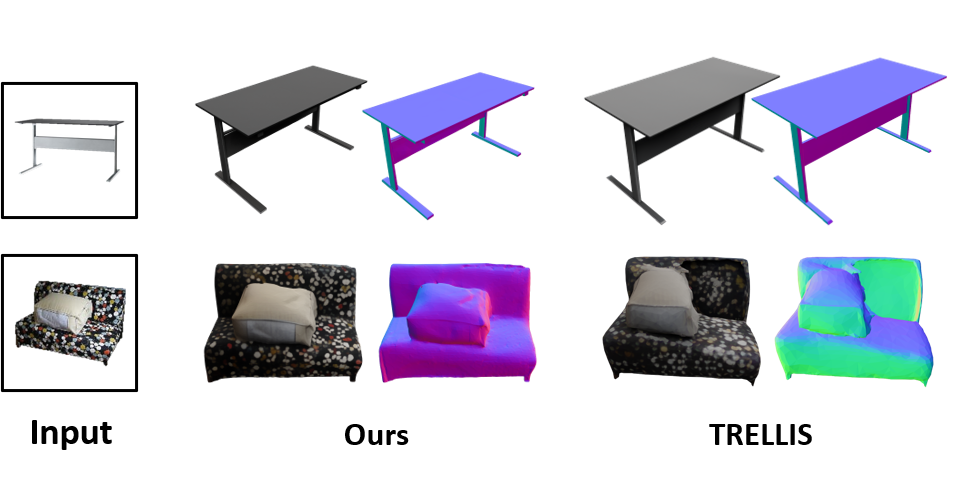}
    \vspace{-4mm}
    \caption{\textbf{Real-world examples on Pix3D.}}
    \vspace{-4mm}
    \label{fig:real-world}
\end{figure}

We further evaluate the robustness of our method on real-world images from the Pix3D~\cite{pix3d} dataset. As illustrated in Fig.~\ref{fig:real-world}, our approach maintains robust performance on real image inputs, producing geometry that aligns more faithfully with the input images compared to the baseline method.

\section{Conclusion}
\label{sec:conclu}
We introduce Points-to-3D, a diffusion-based framework that first leverages explicit 3D point cloud priors as input to enable geometry-controllable 3D asset and scene generation. Built upon the latent 3D diffusion model TRELLIS~\cite{xiang2025structured}, we investigate a natural way to embed point cloud as initialization within the framework. After training TRELLIS’s structure generation network to acquire inpainting capabilities, we employ a staged sampling strategy—structural inpainting followed by boundary refinement—that reconstructs the global geometry while preserving the input visible regions. Experiments demonstrate the benefits of explicitly embedding 3D priors, highlighting a promising direction for controllable and reliable 3D generation in real-world applications.

{
    \small
    \bibliographystyle{ieeenat_fullname}
    \bibliography{main}
}

\clearpage
\setcounter{page}{1}

\appendix

\section{Experimental Details}

Our training dataset consists of object collections from the 3D-FUTURE~\cite{fu20203dfuture} (9,472 objects), HSSD~\cite{khanna2024hssd} (6,670 objects), and ABO~\cite{collins2022abo} (4,485 objects) datasets. For each object, we render the image of the $T = 24$ views, together with the corresponding depth map, and extract the visible point cloud for each view by enforcing depth consistency with a threshold $\tau$ = 0.05 times the depth range (maximum minus minimum depth) in that view. The visible point cloud is then converted into an initial SS latent, which is paired with the original SS latent as ground truth to train the sparse structure flow transformer for inpainting.

For evaluation, we use randomly sampled subset of the Toys4K~\cite{stojanov21toys4k} (500 objects) dataset and 3D-FRONT~\cite{fu20213dfront} (500 scenes) dataset. For each test object or scene, we render 8 views using cameras with yaw angles $(0^\circ, 45^\circ, 90^\circ, 135^\circ, 180^\circ, 225^\circ, 270^\circ, 315^\circ)$ and a fixed pitch angle of $30^\circ$. The camera is positioned at a radius of 1.8 from the object center. For PSNR, SSIM, and LPIPS~\cite{LPIPS}, we directly compare the rendered images of generated results with the rendered images of the ground-truth objects and report the average scores. For the DINO-based similarity metric, we report the average discrepancy between the rendered images of the generated and ground-truth assets, quantified as $(1 - S_{\text{DINO}})$, where $S_{\text{DINO}}$ denotes the DINO similarity score. For the normal-based metric, we render normal maps from the 8 views and compute the average score between the normal maps of the generated and ground-truth assets. For Chamfer Distance (CD) and F-score, we normalize all the objects within the range (-0.5, 0.5) and set the F-score distance threshold to 0.05. During testing, for the point cloud priors input, we align the point cloud to the orientation of the corresponding ground-truth object to ensure that the generation conditioned on this point cloud can be directly evaluated.

\section{More Results}
We provide additional qualitative examples and experimental results to further demonstrate the performance of our method.

\begin{figure}
    \centering
    \includegraphics[width=1\linewidth]{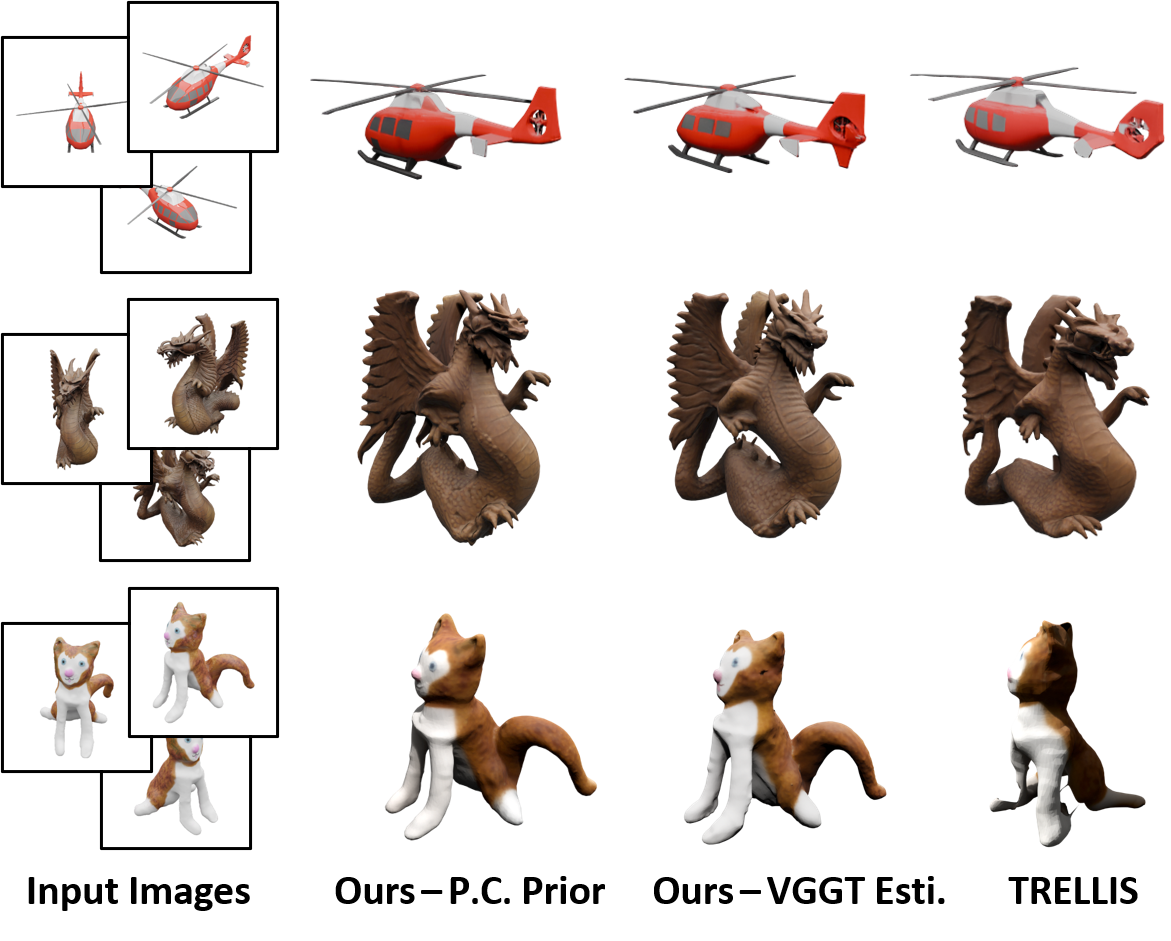}
    \vspace{-12pt}
    \caption{\textbf{Generation results with 3 input views on Toys4K.} The first column of our results uses sampled point-cloud priors extracted from the visible regions of the three input images, whereas the “VGGT-estimated” results rely on point clouds inferred from the input images by VGGT.}
    \label{fig:3view}
\end{figure}

\begin{figure*}
    \centering
    \includegraphics[width=0.8\linewidth]{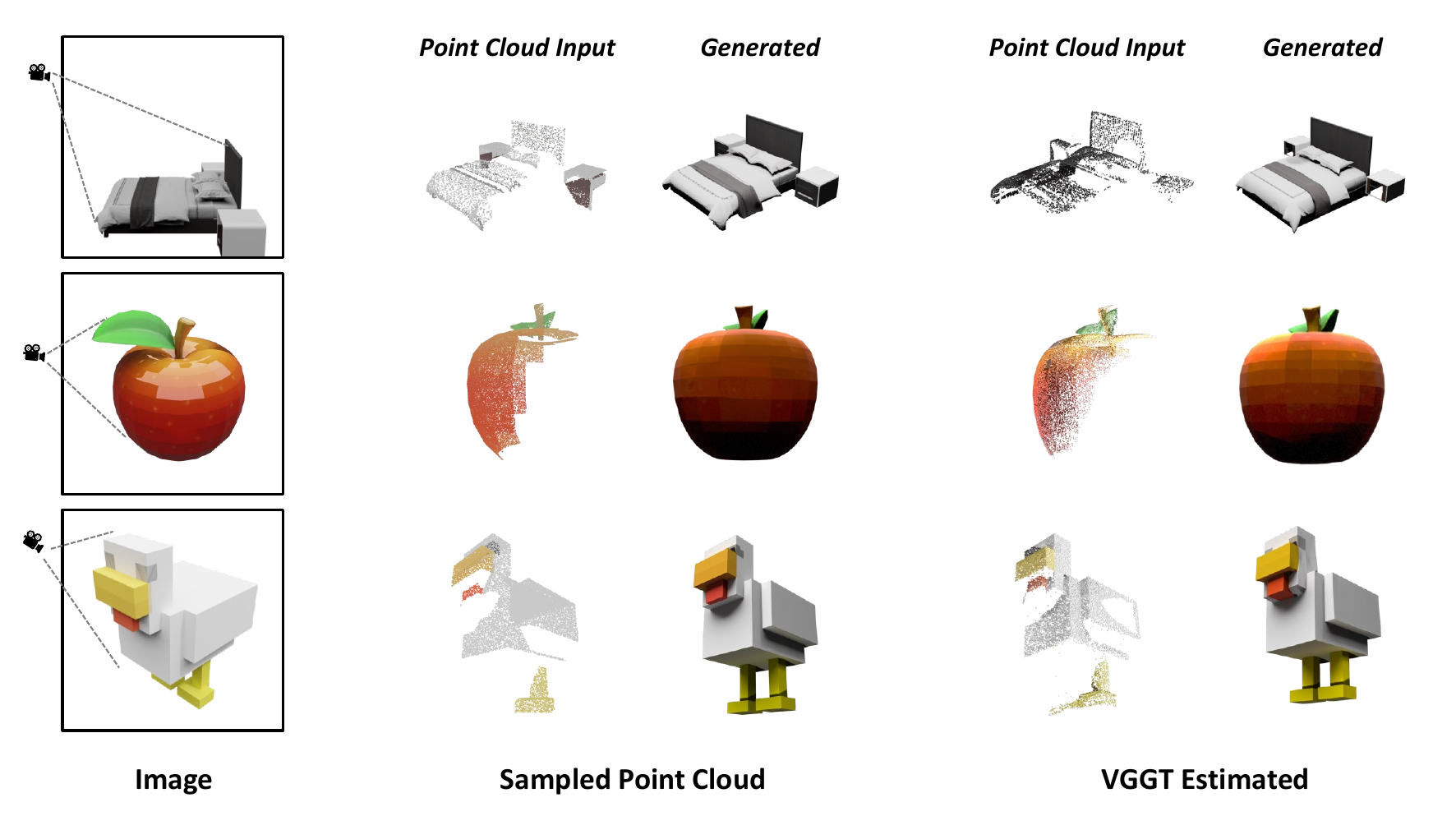}
    \vspace{-2mm}
    \caption{\textbf{Input point cloud priors examples.} We show the observable point cloud priors examples for the two input modes with single-view input in this paper, along with their corresponding generation results.}
    \label{fig:pc_demo}
\end{figure*}
\begin{figure*}
    \centering
    \includegraphics[width=0.96\linewidth]{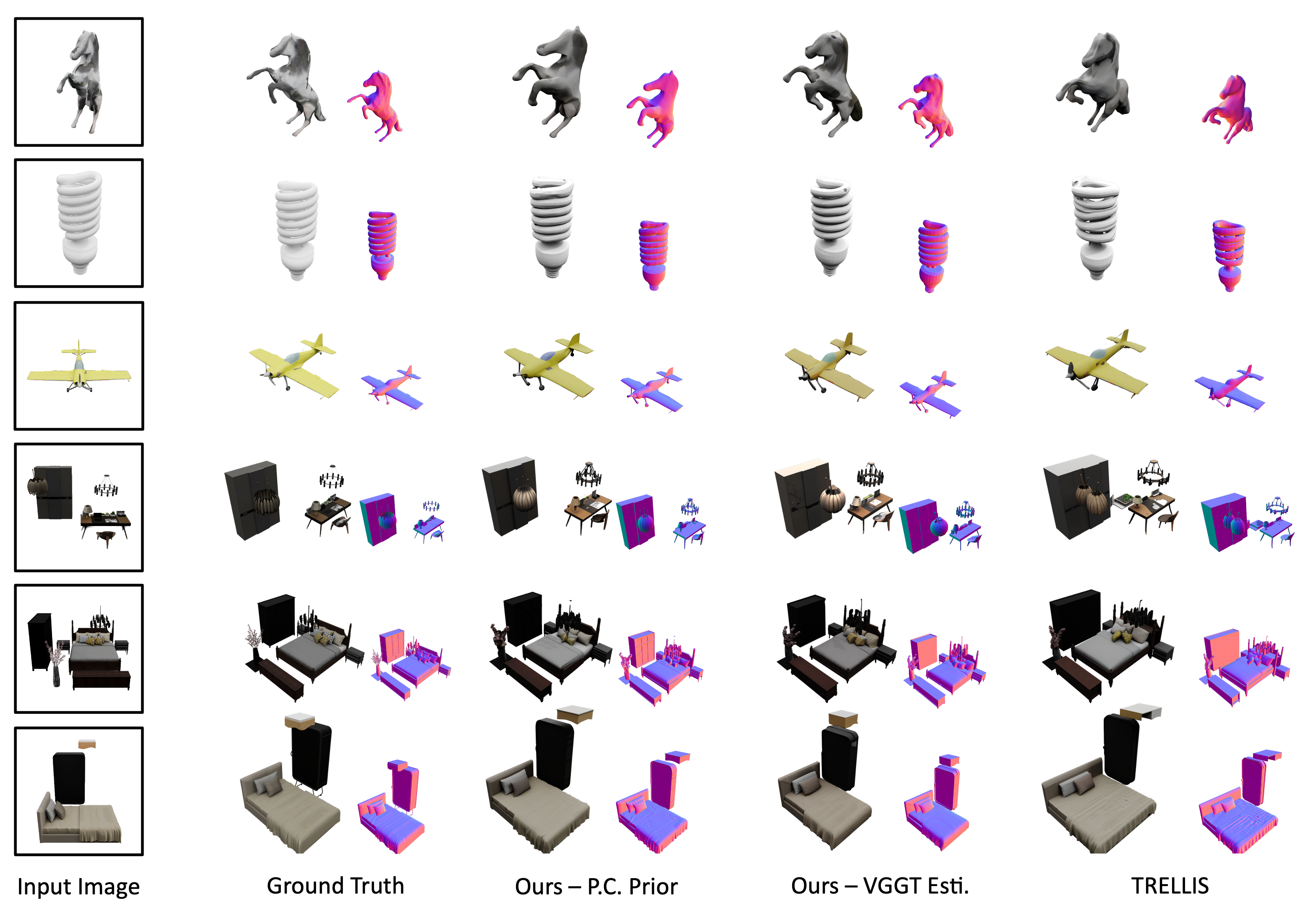}
    \vspace{-2mm}
    \caption{\textbf{More image-to-3D examples.} More single-image to 3D generation visualization results on Toy4K (row 1-3) and 3D-Front dataset (row 4-6).}
    \label{fig:imgto3d}
\end{figure*}

\begin{table*}[t]
  \centering
  \caption{\textbf{Comparison on single-object generation with 3 views input on Toy4K dataset.} 
  }
  \label{tab:toys4k_multi}
  \resizebox{0.83\linewidth}{!}{
  \begin{tabular}{@{}l|cccc|cccc@{}}
    \toprule
     &\multicolumn{4}{c|}{Rendering}  & \multicolumn{4}{c}{Geometry} \\
    Method  & $\text{PSNR}\uparrow$ & $\text{SSIM}(\%)\uparrow$ &  $\text{LPIPS}\downarrow$ & $\text{DINO}(\%)\downarrow$ & $~\text{CD}\downarrow$ & $\text{F-Score}\uparrow$ & $\text{PSNR-N}\uparrow$ & $\text{LPIPS-N}\downarrow$ \\
    \midrule
    TRELLIS~\cite{xiang2025structured}  & 23.19 & 92.63 & 0.075 & 5.79 & ~0.025~ & ~0.904 & 26.22 & 0.066 \\
    \textbf{\mname} (Ours-VGGT Esti.) & 23.44 & 93.21 & 0.057  & 5.58 & ~0.015~ & ~0.971 & 28.35  & 0.035  \\
    \textbf{\mname} (Ours-P.C.Priors)  & \textbf{23.98} & \textbf{94.02} & \textbf{0.050}  & \textbf{5.26} & ~\textbf{0.009}~  & ~\textbf{0.988} & \textbf{30.45}  & \textbf{0.028} \\
    \bottomrule
  \end{tabular}}
\end{table*}

\subsection{Multi-Views Input Generation}
Because our flow-based model performs iterative denoising, it can directly incorporate multi-view reference images as conditioning inputs at different denoising steps. For VGGT-estimated point clouds, multi-view inputs produce more accurate predictions; and greater point cloud coverage consistently leads to better reconstruction. We further evaluate the case of using three input views on Toys4K~\cite{stojanov21toys4k} dataset. Specifically, we first feed the multi-view reference images into VGGT~\cite{wang2025vggt} to obtain a more complete predicted point cloud. As shown in Tab.~\ref{tab:toys4k_multi}, while multi-view input naturally improves the baseline TRELLIS~\cite{xiang2025structured} geometry, our method achieves substantially higher structural accuracy, consistently maintaining controllable geometry. For accurate point cloud priors, we extract the visible sampled surface point cloud from the three views using depth consistency and use it as the input prior. With these priors, our method produces reconstructions that are very close to the ground truth. Fig.~\ref{fig:3view} further shows the visualization comparisons. These results demonstrate the robustness and effectiveness of our method across different numbers of input images.

\begin{figure}
    \centering
    \includegraphics[width=1\linewidth]{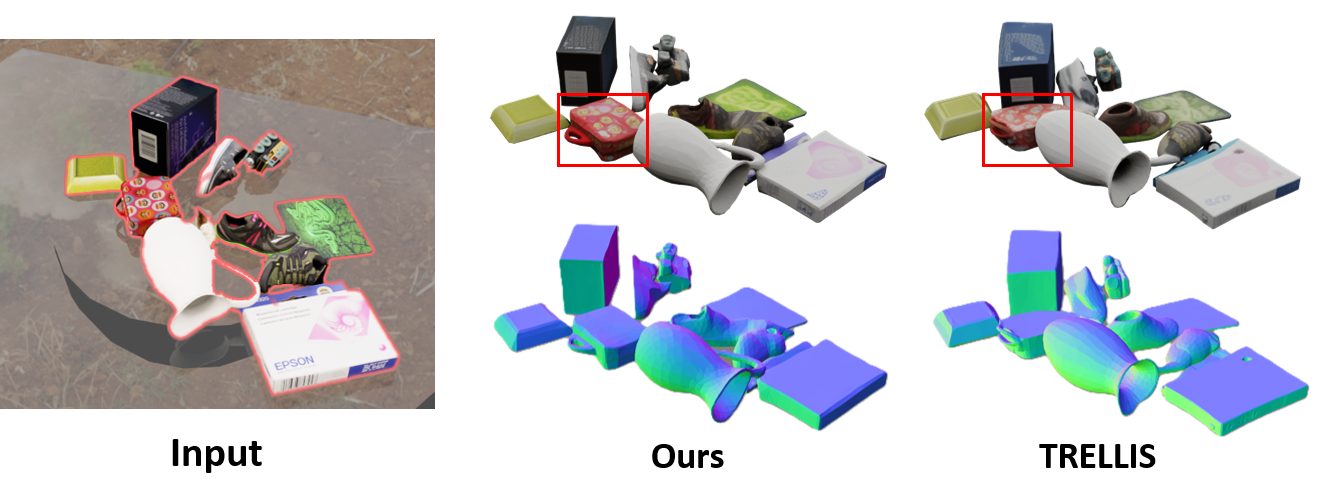}
    \caption{Example of cluttered table with depth sensor.}
    \label{fig:table}
\end{figure}

\begin{figure}
    \centering
    \includegraphics[width=1\linewidth]{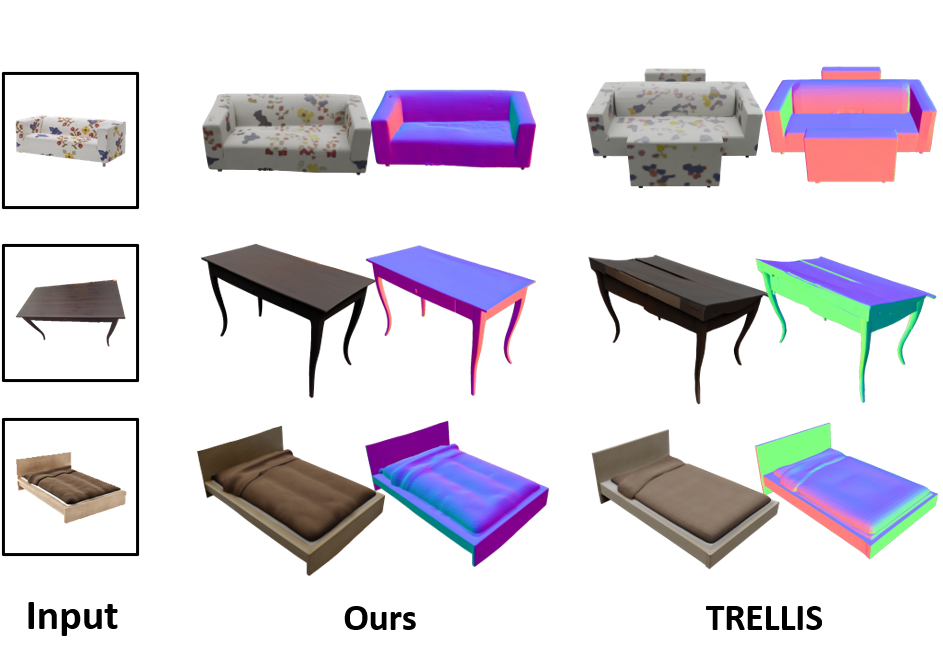}
    \caption{\textbf{More real-world image generation examples.}}
    \label{fig:realworlda}
\end{figure}

\begin{figure}
    \centering
    \includegraphics[width=1\linewidth]{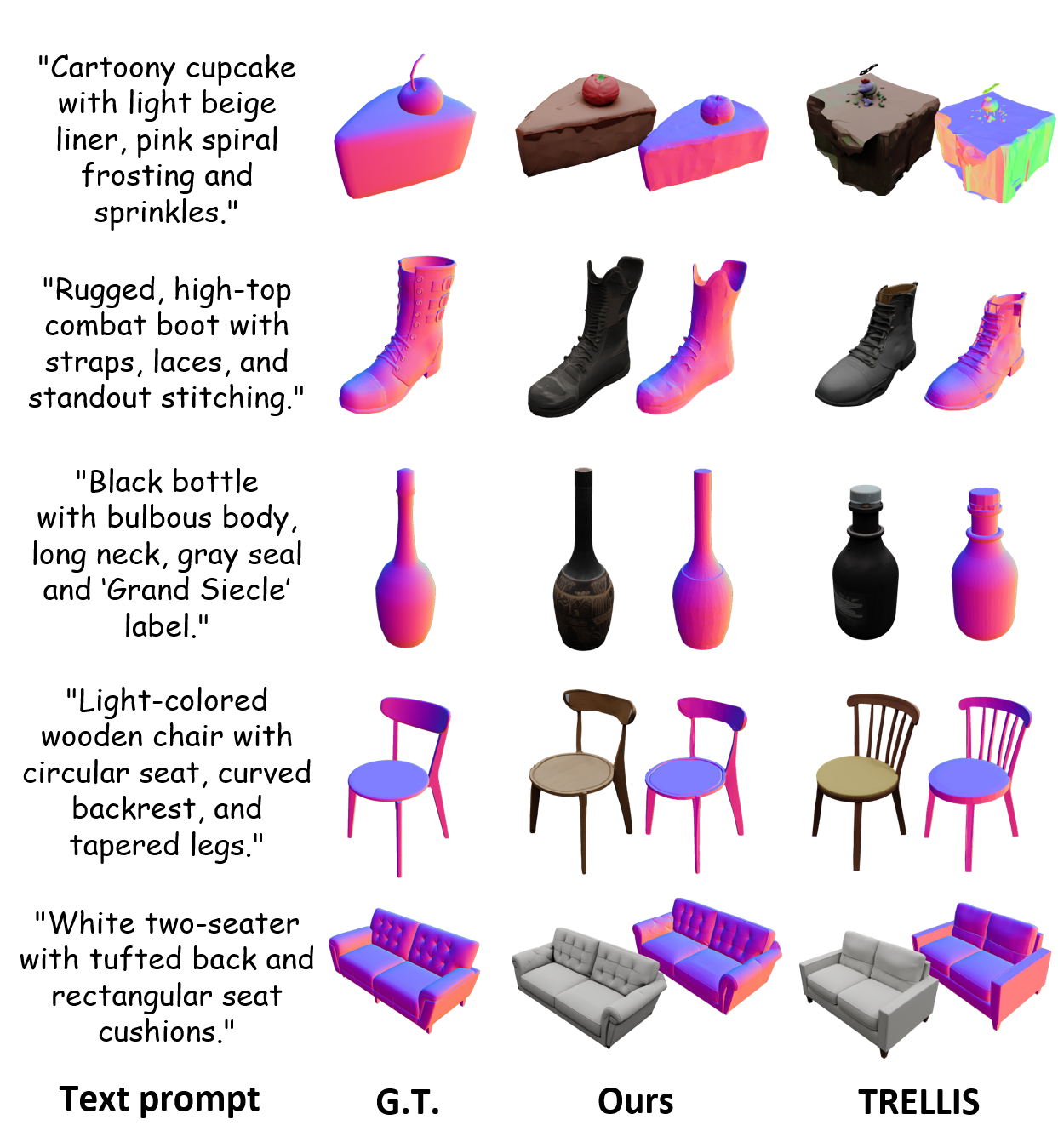}
    \caption{\textbf{Text-to-3D generation examples.}}
    \label{fig:txta}
\end{figure}

\begin{table}[t]
  \centering
  \caption{\textbf{Comparison of text-to-3D generation on Toys4K.} 
  }
  \label{tab:txt}
  \resizebox{\linewidth}{!}{
  \begin{tabular}{l|ccccc}
    \toprule
    Methods & $\text{CLIP}\uparrow$  & $\text{CD}\downarrow$ & $\text{F-Score}\uparrow$ & $\text{PSNR-N}\uparrow$ & $\text{LPIPS-N}\downarrow$ \\
    \midrule
    LGM~\cite{tang2024lgm} & 0.247 & 0.086 & 0.412 & 19.55 & 0.223 \\
    TRELLIS~\cite{xiang2025structured} & 0.298 & 0.047 & 0.639 & 21.25 & 0.159 \\
    \textbf{\mname} & \textbf{0.299} & \textbf{0.022} & \textbf{0.892} & \textbf{24.75} & \textbf{0.094} \\
    \bottomrule
  \end{tabular}}
\end{table}

\subsection{Point Cloud Priors Examples}
In Fig.~\ref{fig:pc_demo}, we illustrate examples of the two types of point cloud priors considered in this work, which correspond to the two most common practical scenarios: (1) partial point clouds directly captured by hardware sensors (e.g., LiDAR on an iPhone), and (2) point cloud estimated from input images via feed-forward point-map prediction (e.g., VGGT~\cite{wang2025vggt}). This experimental setup enables a comprehensive evaluation of our method over a broader spectrum of practical cases. As shown in Fig.~\ref{fig:pc_demo}, these visible-region priors impose reliable geometric constraints that steer our model toward controllable and faithful 3D generation.

\subsection{More Image-to-3D Examples}
We provide additional visualization results for image-to-3D generation in Fig.~\ref{fig:imgto3d}, demonstrating the effectiveness of our method. Experiments highlight that our method addresses a major limitation of existing 3D generation frameworks that struggle to fully incorporate available 3D information, and achieves substantial improvements in both single-object and scene-level generation.
In Fig.~\ref{fig:table} we also tested our model on cluttered table in RaySt3R~\cite{duisterhof2025rayst3r} dataset with depth sensors and found it shows capability in handling such complex cases. Compared to TRELLIS, our method better preserves 3D geometry under this extreme out-of-distribution condition, highlighting the advantage and robustness of our method. 

\subsection{More Real-world and Text-to-3D Examples}
We showcase more results in real-world image generation in Fig.~\ref{fig:realworlda}, demonstrating the robustness of our method in practical scenarios. Moreover, we also assess our model under text-to-3D settings on Toys4K~\cite{stojanov21toys4k}, where text prompts and point cloud priors are provided as input. As shown in Tab.~\ref{tab:txt} and Fig.~\ref{fig:txta}, our method successfully generates geometries that are semantically consistent with the input prompts and structurally well-controlled by the given point cloud priors.

\begin{table}[t]
  \centering
  \caption{\textbf{Noisy point cloud priors.} We add different levels of perturbation to the accurate point-cloud priors to evaluate the impact of noisy 3D priors, and present the results of our simple repair process for noisy point cloud inputs below.
  }
  \label{tab:noisey}
  \resizebox{1\linewidth}{!}{
  \begin{tabular}{l|cccc}
    \toprule
    Methods  & $\text{CD}\downarrow$ & $\text{F-Score}\uparrow$ & $\text{PSNR-N}\uparrow$ & $\text{LPIPS-N}\downarrow$  \\
    \midrule
    \textbf{P.C. priors} & 0.013  & 0.964  & 27.10 & 0.053   \\
    + 5\% perturbation & 0.022 & 0.910 & 24.32 & 0.082  \\
    + 10\% perturbation & 0.031 & 0.817 & 22.87 & 0.109  \\
    + 15\% perturbation & 0.045 & 0.667 & 20.36 & 0.181  \\
    \midrule
    + 10\% perturbation \& repair & 0.027 & 0.855 & 23.47 & 0.098  \\
    + 15\% perturbation \& repair & 0.036 & 0.791 & 21.85 & 0.132  \\
    \bottomrule
  \end{tabular}}
\end{table}

\subsection{Noisy Point Clouds Input}
Our method is primarily designed for settings where reliable 3D priors are available, where the goal is not to enhance existing point-cloud quality, but to faithfully preserve and explicitly integrate existing high-quality 3D priors—such as those obtained from LiDAR sensors (now reliable and widely available on smartphones)—into 3D generation frameworks, thereby enabling current 3D generation models to better leverage sensed 3D data and future advances in feed-forward point-map prediction methods. Nevertheless, we observe that low-quality point cloud inputs remain a key limitation of our current implementation, as they can negatively impact the overall performance.

We analyze the effect of noisy point clouds by adding random noise within a fixed error distance range, as shown in Tab.~\ref{tab:noisey}: our model remains robust to mild perturbations but degrades under heavy noise, which is expected since it explicitly relies on the visible prior. Although improving the quality of the priors is not the primary focus of this work, we propose a simple strategy that diffuses the initial noisy structure through a few sampling steps prior to the inpainting process. This design can be seamlessly integrated into our pipeline with no additional cost, leading to improved performance under noisy inputs, as shown in Tab.~\ref{tab:noisey}. However, in our current implementation, we do not recommend fully relying on unreliable 3D priors in the presence of noise. Addressing such cases requires further investigation in future work.

\end{document}